\documentclass[conference]{IEEEtran}
\IEEEoverridecommandlockouts
\usepackage{cite}
\usepackage{amsmath,amssymb,amsfonts}
\usepackage{algorithmic}
\usepackage{graphicx}
\usepackage{textcomp}
\usepackage[dvipsnames]{xcolor}

\usepackage{amsmath}
\usepackage{amsthm}
\usepackage{amsfonts}

\usepackage{booktabs}
\usepackage{algorithm}

\usepackage{multicol}
\usepackage{multirow}
\usepackage{subfig}

\usepackage{tikz}
\usetikzlibrary{backgrounds}
\usetikzlibrary{arrows,shapes}
\usetikzlibrary{tikzmark}
\usetikzlibrary{calc}

\usepackage{amsmath}
\usepackage{amsthm}
\usepackage{amssymb}
\usepackage{mathtools, nccmath}
\usepackage{wrapfig}
\usepackage{comment}
\usepackage{tcolorbox}

\usepackage{tikz}
\usetikzlibrary{arrows,shapes,positioning,shadows,trees,mindmap}
\usepackage[edges]{forest}
\usetikzlibrary{arrows.meta}
\colorlet{linecol}{black!75}
\usepackage{xkcdcolors} % xkcd colors

\usepackage{tikz}
\usetikzlibrary{backgrounds}
\usetikzlibrary{arrows,shapes}
\usetikzlibrary{tikzmark}
\usetikzlibrary{calc}
\newcommand{\highlight}[2]{\colorbox{#1!17}{$\displaystyle #2$}}
\colorlet{mhpurple}{Plum!80}

\renewcommand{\highlight}[2]{\colorbox{#1!17}{#2}}

\def\BibTeX{{\rm B\kern-.05em{\sc i\kern-.025em b}\kern-.08em
		T\kern-.1667em\lower.7ex\hbox{E}\kern-.125emX}}
\begin{document}
	
	\title{Adversarial Deep Reinforcement Learning for Improving the Robustness of Multi-agent Autonomous Driving Policies\\
		%{\footnotesize \textsuperscript{*}Note: Sub-titles are not captured in Xplore and
		%should not be used}
		%\thanks{Identify applicable funding agency here. If none, delete this.}
	}
	\author{\IEEEauthorblockN{Aizaz Sharif}
		\IEEEauthorblockA{\textit{Simula Research Laboratory} \\
			Oslo, Norway \\
			aizaz@simula.no}
		\and
		\IEEEauthorblockN{Dusica Marijan}
		\IEEEauthorblockA{\textit{Simula Research Laboratory} \\
			Oslo, Norway \\
			dusica@simula.no}
		
	}
	%\and
	%\IEEEauthorblockN{3\textsuperscript{rd} Given Name Surname}
	%\IEEEauthorblockA{\textit{dept. name of organization (of Aff.)} \\
	%\textit{name of organization (of Aff.)}\\
	%City, Country \\
	%email address or ORCID}
	%\and
	%\IEEEauthorblockN{4\textsuperscript{th} Given Name Surname}
	%\IEEEauthorblockA{\textit{dept. name of organization (of Aff.)} \\
	%\textit{name of organization (of Aff.)}\\
	%City, Country \\
	%email address or ORCID}
	%\and
	%\IEEEauthorblockN{5\textsuperscript{th} Given Name Surname}
	%\IEEEauthorblockA{\textit{dept. name of organization (of Aff.)} \\
	%\textit{name of organization (of Aff.)}\\
	%City, Country \\
	%email address or ORCID}
	%\and
	%\IEEEauthorblockN{6\textsuperscript{th} Given Name Surname}
	%\IEEEauthorblockA{\textit{dept. name of organization (of Aff.)} \\
	%\textit{name of organization (of Aff.)}\\
	%City, Country \\
	%email address or ORCID}

	\maketitle
	
	\begin{abstract}
		%Deep reinforcement learning is widely used to train autonomous cars in a simulated environment. Still, autonomous cars are well known for being vulnerable when exposed to adversarial attacks. This raises the question of whether we can train the adversary as a driving agent for finding failure scenarios in autonomous cars, and then retrain autonomous cars with new adversarial inputs to improve their robustness. In this work, we first train and compare adversarial car policy on two custom reward functions to test the driving control decision of autonomous cars in a multi-agent setting. Second, we verify that adversarial examples can be used not only for finding unwanted autonomous driving behavior, but also for helping autonomous driving cars in improving their deep reinforcement learning policies. By using a high fidelity urban driving simulation environment and vision-based driving agents, we demonstrate that the autonomous cars retrained using the adversary player noticeably increase the performance of their driving policies in terms of reducing collision and off-road steering errors.
		%Deep reinforcement learning is widely used to train autonomous cars in a simulated environment. Still, 
  Autonomous cars are well known for being vulnerable to adversarial attacks that can compromise the safety of the car and pose danger to other road users. To effectively defend against adversaries, it is required to not only test autonomous cars for finding driving errors but to improve the robustness of the cars to these errors. To this end, in this paper, we propose a two-step methodology for autonomous cars that consists of (i) finding failure states in autonomous cars by training the adversarial driving agent, and (ii) improving the robustness of autonomous cars by retraining them with effective adversarial inputs. Our methodology supports testing autonomous cars in a multi-agent environment, where we train and compare adversarial car policy on two custom reward functions to test the driving control decision of autonomous cars. We run experiments in a vision-based high-fidelity urban driving simulated environment. Our results show that adversarial testing can be used for finding erroneous autonomous driving behavior, followed by adversarial training for improving the robustness of deep reinforcement learning-based autonomous driving policies. We demonstrate that the autonomous cars retrained using the effective adversarial inputs noticeably increase the performance of their driving policies in terms of reduced collision and offroad steering errors.
	\end{abstract}
	
	
	\section{Introduction}
	Autonomous cars (ACs, also known as self-driving cars) are complex technologies that are prone to failures~\cite{Garcia}. ACs integrate deep learning based software, which is known to be difficult to validate~\cite{Riccio}\cite{MLTest}; yet, testing and validating ACs is indispensable for their deployment in practice ~\cite{R49}\cite{koopman2016challenges}\cite{7823109}. While there has been progress made by researchers on testing AI-based models~\cite{R5}\cite{R23}, testing of ACs is another complex area to tackle, due to several reasons. 
	
	First, a lot of research has been proposed on scenario and test case generation~\cite{li2020av}\cite{8917103}\cite{Alessio}\cite{R7}\cite{R32} and input validation~\cite{DeepRoad}\cite{DeepTest}\cite{R44}\cite{R2}\cite{R11} for testing and validating AC driving models. While such approaches can expose failures in ACs, they are only focused on error \textit{detection} and not \textit{correction}. There is limited research work on %On the other hand, there are very limited examples in the validation and verification of ACs research where we go one step ahead of testing techniques and use the analysis of 
	analyzing the failed AC driving systems for understanding out-of-distribution scenes and edge cases that need to be induced in the training of AC models. Therefore, there is a need for a comprehensive AC testing methodology able to not only discover errors in AC driving models but also %their driving performance is studied and 
	improve the performance of failing AC models given the same attacking inputs. This would help in overcoming existing errors and improving robustness in the evolving AI-based AC systems.
	
	\begin{figure}[!htbp]
		\centering
		
		\includegraphics[width = 0.49\textwidth,height=0.10\textheight]{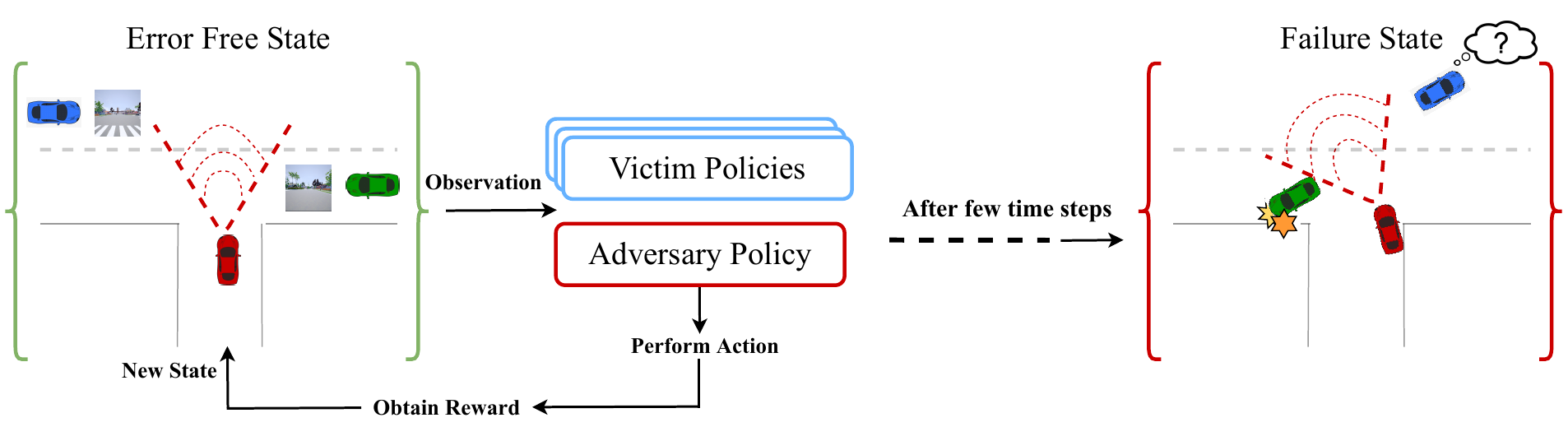}
		\caption{Illustration of the first step of the MAD-ARL framework, where adversary is driving AC under test into failure states.}	\label{fig:First}
	\end{figure}

	Second, existing research efforts often address AC testing in simplistic evaluation scenarios. For example, AC testing is often considered as a \textit{single-agent} problem, where only one car is taken as a system under test (SUT)~\cite{li2020av}\cite{majumdar2021paracosm}\cite{R1}\cite{R2}\cite{R11}. While a single-agent self-driving environment still has open challenges, there is a need to advance and test a multi-agent self-driving, as it represents a more realistic environment. In the near future, multiple ACs will co-exist on the road, and the more such cars start interacting with each other, as well as with human drivers, the more complex their testing becomes~\cite{R13}. 
	%Multi-agent systems have been studied extensively in the field of reinforcement learning (RL) by various research groups~\cite{R6}\cite{R22}\cite{R25}, but they are still behind to be deployed in the AC domain on a larger scale, partly due to the lack of proper testing. 
	Another type of simplistic scenario is validating ACs in \textit{lane keeping}~\cite{Alessio}\cite{R34} and \textit{mixed traffic systems}~\cite{R4} simulated environments. Existing industrial-grade AC testing uses vision-based end-to-end driving systems where ACs are tested on the high dimensional stream of inputs in a partially observable environment, unlike in lane-keeping simulated environments where such scenarios rely on low dimensional inputs and a fully observable environment for making driving decisions. Moreover, multi-agent driving environments consider cooperative and dependent driving agents, while the majority of AC research considers driving agents as independent and non-communicating competitive agents. Recent work~\cite{R4} made progress toward testing connected and automated vehicles by injecting adversarial noise in a mixed-traffic-based network environment. However, this work is of limited practical use as it does not consider multiple non-communicating AC agents.
	Another example of a simplistic evaluation scenario is testing of \textit{rule-based} driving systems in a simulated environment~\cite{R8}\cite{R28}. Such driving systems are based on deterministic logic and do not scale to corner cases, unlike ACs based on deep learning systems.
	%is constantly evolving for the autonomous driving (AD) community, but also due to its probabilistic nature, it is very complex to test them, unlike rule-based AI models which are based on deterministic logics and do not scale to corner cases.
	One more type of simplistic evaluation scenario is \textit{offline testing} of deep learning based AC models~\cite{9159088}. Offline testing involves validating the performance and control decisions of a vision-based AC model by only using datasets of urban simulated driving scenarios, collected by humans. Even though offline testing contributes to the machine learning and autonomous driving (AD) community, it is very restricted to a single-agent setting (offline testing). Adding multi-agent actions induced by other driving policies in a non-stationary environment~\cite{Papoudakis2019DealingWN} is a very crucial and new area to tackle in the AC driving and testing community. 

Third, deep reinforcement learning (DRL) algorithms are extensively used in training vision-based safe AC models in urban driving environments~\cite{PPO2}\cite{A3C2}\cite{DQN2}\cite{DDPG}\cite{DDPG2}. One way to test their driving behavior is using adversarial RL (ARL) since DRL is proven to be vulnerable multiple times against adversarial attacks. Existing research suggests that ARL-based agents can be effective in exposing vulnerabilities of DRL-based agents in a blackbox manner~\cite{R6}. However, the idea has been explored in a simplistic driving environment~\cite{R8}\cite{R4}. In our work, we make use of ARL for discovering effective attacking inputs that we further use to improve the robustness of DRL-based AC policies in a complex non-communicating vision-based urban driving environment. Specifically, we introduce ARL as part of a driving simulation in order to add adversarial actions against the AC policies under test. By doing so, we show not only find failure scenarios of the DRL-based ACs interacting with the adversarial drivers but to leverage effective adversarial actions to improve the AC driving robustness. 
	
	%Since our target ACs are also trained on the same RL algorithms, we want to further investigate the use of ARL as a driving agent
	%
	% as a blackbox testing mechanism
	
	% In the non-AC testing community recent findings in~\cite{R6} advances in adversarial RL as a blackbox testing to expose more vulnerabilities of RL-based AC agents in a two-player game. 
	%
	%Since the same RL algorithms are extensively used in training safe vision-based AC models, there is a lack of work on blackbox adversarial testing of ACs driving environment.

	%While there is an extensive research on having whitebox testing, recent findings in~\cite{R6} advances in adversarial RL to expose vulnerability of RL-based agents in a two-player game, where adversary is able to confuse victim agent without any whitebox access to the victim policy. 
	%
	%single-agent ACs, by finding failure test cases~\cite{R7}\cite{R32}.  
	%RL on the other hand has been used in different styles to test autonomous driving within simulations~\cite{R8}\cite{R34}.  
	
	To address these three challenges, in this paper we propose a framework \textbf{\lq}\textbf{M}ulti-\textbf{A}gent \textbf{D}riving with \textbf{A}dversarial \textbf{R}einforcement \textbf{L}earning \textbf{MAD-ARL}\textbf{\rq} which is a novel approach for improving the robustness of non-communicating multi-agent ACs using an adversarial agent in an urban driving scenario. In the first step, we train an adversarial agent car that aims to create \textit{natural observations} that are adversarial for the ACs under test. Using the trained adversary, we test multi-agent policies of ACs under test, with the goal of exposing faults in the cars' driving policies as illustrated in Figure~\ref{fig:First}. In the second step, 
	%The figure shown the first step of our testing framework. By training adversarial agent car in a shared driving environment against one victim AC, we test multi-agent victim policies to creates \textit{natural observations} that are adversarial for the victims.	Such adversarial actions drives them into error states after number of simulation steps.  Then 
	we retrain the ACs under test with the adversarial policies to defend against adversarial attacks. The results are compared with the baseline autonomous models (adversary-free trained policies) to evaluate the effectiveness of the retraining strategy as a defense against adversarial attacks. In our experiments, we demonstrate that a trained adversarial player can improve the robustness of more than one vision-based AC policy in terms of fewer collisions and offroad steering accidents. The main idea is to use adversarial examples beyond testing purposes to improve the robustness of ACs, since adversarial attacks are usually not considered when AC models are being trained in urban driving scenarios.
	%The same adversarial agent will be used to improve the robustness of more than one victim driving car.
	
	The key research contributions in this paper are:
	\begin{enumerate}
		% New changes
		\item Proposing a two-step methodology for finding failure scenarios in ACs and improving the robustness of ACs given the effective test scenarios.
		
		\item Introducing a novel DRL framework for testing and improving driving policies of independent non-communicating agents in a multi-agent AC environment. The implementation of the framework is open~\footnote{\label{https://github.com/T3AS/MAD-ARL} https://github.com/T3AS/MAD-ARL} and reusable, which supports the reproducibility of research in the AD domain.
		
		\item Designing an RL-based adversarial agent that can be generalized for testing more than one AC driving policy by only training against a single victim AC.
		
		\item Designing an adversary that can effectively drive an AC into error states by creating natural (i.e. realistic) observations for the AC's driving policies, without whitebox access to its input state.
		
		\item Experimentally demonstrating that retraining DRL-based AC driving policies using adversarial driving models can be an effective defense against adversarial attacks. 
%		The experimental results are provided in an anonymous repository~\textsuperscript{\ref{https://github.com/T3AS/MAD-ARL}}.

		%----------------------------------------------------
		%	\item Introducing a deep RL framework for testing driving policies of independent non-communicating agents in a multi-agent AC environment.
		%	
		%	\item Proposing an RL-based adversarial agent that can be generalized for testing more than one AC driving policies by only training against a single victim AC.
		%	
		%	\item Training an adversary that can effectively drive an AC into error states by creating natural (i.e. realistic) observations for the AC's driving policies, without whitebox access to its input state.
		%
		%	
		%	\item Experimentally demonstrating that retraining deep RL-based driving policies of ACs using adversary driving models can be an effective defense against adversarial attacks.

	\end{enumerate}
	
	%%%%%%%%%%%%%%%%%%%%%%%%%%%%%%%%%%%%%%%%%%%%%%%%%%%%%%%%%%%%%%%%%%%%%%%%
	
\section{Related Work}\label{sec:Related_Work}

The majority of related work has focused on generating test scenarios for discovering errors in ACs and adversarial testing of ACs. The main limitations of these works are the lack of focus on improving the robustness of ACs once errors are discovered, as well as simplistic evaluation conditions. Next, we summarize the main approaches, discussing their benefits and limitations.    

\subsection{Test Scenario Generation for AC}
Authors in~\cite{DeepRoad} use GANs to generate synthetic images to validate the driving robustness of deep learning-based autonomous driving systems. They also use metamorphic testing to check the consistency of the model outputs against different types of synthetic images. Another close work~\cite{DeepTest} proposes a systematic testing tool for evaluating DNN-based AC models. They do so by generating test cases using real-world conditions such as rain and lightning conditions. They perform DNN logic coverage by adding transformations to test inputs within Udacity self-driving car challenge simulator~\cite{chauffeur}. Similar to~\cite{DeepRoad}, they also use domain-specific metamorphic relations to find fault behaviors of DNN. While the proposed work has achieved great results, it is limited in way that the driving scenarios are only tested in a single-agent environment. Also, both~\cite{DeepTest} and~\cite{DeepRoad} would benefit from adding the same synthetic images in the retraining of the tested DNN AC models to compare the robustness with the baseline models.

The authors in~\cite{Alessio} use a search-based testing technique to automatically create challenging virtual scenarios for testing self-driving cars. These scenarios are used to test AI driving models such as DeepDriving~\cite{chen2015deepdriving} to perform systematic testing of lane-keeping systems. While the work contributes to having more complex evaluation scenarios, it does not address the problem of AC testing in a realistic multi-agent driving environment. Authors in~\cite{8917103} use Bayesian optimization for test case generation for ACs. The proposed work learns parameters using the system's output to create test case scenarios that lead AC into failure states. While the authors are able to identify test cases for complex black boxes like autonomous systems, the work lacks complex driving scenarios with more than one AC in the same environment for testing purposes. 
	
A thorough case study by authors in~\cite{9159088} performs a comparative study of the pros and cons of testing deep learning-based AC models in offline datasets versus online simulation testing. Offline testing focuses on prediction errors against the dataset, while online testing looks for safety violations within driving scenarios. The tests are performed using a pretrained Udacity car simulator driving model. As a limitation, the work needs an extension of multi-agent testing configurations within online and offline driving scenarios to observe which method will be more beneficial for multi-agent ACs testing. Authors in~\cite{klischat2020scenario} use OpenStreetMap traffic simulator SUMO to suggest a workflow for generating a collection of challenging and safety-critical test scenarios for the safety validation of motion planning algorithms in automated vehicles. As a limitation, the work requires multi-agent AC testing to take advantage of the publicly available generated scenarios.

Authors in~\cite{li2020av} propose an automated fuzzing framework to produce AC safety violation driving scenarios. Using the industrial-grade autonomous driving platform Baidu Apollo, they use domain knowledge of vehicle dynamics and genetic algorithms to find failure scenarios. The experiments are performed in a partial multi-agent environment with one AC under test driving alongside non-AC traffic, as non-players. As a limitation, the work does not address the problem of improving the robustness of the same ACs using industrial-grade urban driving simulators. Furthermore, the work could benefit from adding more than one AI-based AC for testing. Another work in~\cite{majumdar2021paracosm} proposes a programmatic interface that enables designing parameterized environments and test cases for ACs. These test parameters control the behavior and positioning of various actors alongside AC under test, and support test input generation strategies. While the experiments are performed by training the neural network-based driving models, the work is limited to a single-agent AC environment. Also, the work could benefit from parameterized environments for generating edge cases that can increase the robustness of ACs. Furthermore, authors in~\cite{deepxplore} propose a whitebox method for testing ACs by triggering as many neurons in the driving model as possible for finding failure scenarios. They pose an optimization problem and apply gradient ascent over the results of test inputs in order to maximize the chance of finding corner cases. As a limitation, the driving scenarios are only tested in a single-agent environment. In contrast, our work is focused on blackbox adversarial testing of ACs in a multi-agent driving environment.

\subsection{Adversarial Testing of ACs}
Recent work~\cite{R4} proposes to use RL-based driving agents to test connected cars by perturbing both the inputs and outputs of a car during training. However, this approach targets mixed-traffic driving with a single AC and multiple human-driven cars, thus it does not consider complex scenarios having more than one non-communicating AC agent.
Another work~\cite{R8} performs adversarial RL for testing a multi-agent driving environment by training more than one adversarial RL agent against one rule-based driving model. While the results look promising, the approach only covers the cases where the trained adversarial cars are exposed to a single non-AI model. As another limitation, the approach has not been evaluated on more complex adversarial driving scenarios, such as T-intersection, which we target in our work.

Another work~\cite{R34} uses RL to stress-test ACs in a simulated environment. The extension of this work~\cite{R28} proposes the idea of reward augmentation for increasing the search space and also finding failure cases in driving policies. Compared to our work, they lack multi-agent test cases even on a small scale. Besides, the work is tested neither in a vision-based simulator nor in real-world driving conditions. Furthermore, while the work improves driving conditions for experiments, it uses adversarial perturbations as noise in the simulation model itself. In contrast, our work adds perturbations by the adversarial car's policy, thus adding adversarial actions as example trajectories for improving AC's driving policies.

Authors in~\cite{R1} proposes a Bayesian optimization-based method for testing ACs. Their method involves creating adversarial scenarios in a Carla-based urban driving simulation~\cite{R51} to expose the weaknesses of autonomous driving policy. Another work~\cite{R2}\cite{R11} also uses an optimization technique for producing physical attacks on driving lanes, in order to attack vision-based driving models. Compared to our work, these works are lacking multi-agent AC scenarios. Authors in~\cite{R32} use a GAN model to generate adversarial objects able to attack LiDAR-based driving systems. Another work~\cite{R35} uses GAN to apply metamorphic testing to CNN-based driving models. Authors in~\cite{PALM} propose a stress testing methodology for LiDAR based perception. Using a real-world driving dataset, they use various weather conditions to test the performance of autonomous driving systems. However, neither of these approaches has been tested in an RL-based multi-agent AC environment. 

	\section{MAD-ARL Formulation}\label{sec:MAD-ARL}

	Our work addresses the problem of adversarial testing of autonomous cars in a multi-agent driving environment for the purpose of (\textbf{i}) finding failures in AC driving models, and (\textbf{ii}) improving the robustness of AC driving models against these failures. 
	
	We model our problem as a 2-player Markov game~\cite{R56}, where one type of a player is the autonomous driving agent under test, which we call a \textit{\textbf{victim}}, and the other player is the adversarial driving agent, which we call an \textit{\textbf{adversary}}, and who is trying to exploit the weakness of the victim. We denote our victims and adversary agent as $T_1$, $T_2$ and $\alpha$, respectively (we consider two ACs under test). The Markov game $M=(S,O,(A_{T1},A_{T2},A_\alpha),P,(R_{T1},R_{T2},R_\alpha))$ in a multi-agent environment consists of $O$ set of state observations and $A_{T}$ $A_\alpha$ represents action set. $P$ is a joint state transition probability function $P : S \times A_{T} \times A_\alpha \mapsto \bigtriangleup (S)$, where $\bigtriangleup (S)$ defines the probability distribution of the next state. Reward function $R$ is based on maximizing the cumulative sum of rewards as $R: S \times A_{T} \times A_\alpha \mapsto \mathbb{R}$. Each player in the set $\{T_1, T_2, \alpha\}$ depends on the current state observation to perform actions and reach the next state while receiving the desired rewards. 
	
	\subsection{Finding Failures in AC Driving Policies}
	The adversary and victim agents work as \textit{\textbf{independent non-communicating competitive}} players. This means that they have no white box access to each other's input state, as well as no shared information to weights parameters. The victim agents are first given the shared environment to train their policies $\pi_{T1}$ and $\pi_{T2}$ in the absence of an adversarial player. The adversary, however, is provided access to the action state sampled from $\pi_T$. %, which is the policy of the self-driving players.
	Since the adversary's policy $\pi_\alpha$ is trained using pre-trained AC policies, we assume that the victim players have fixed weights during adversarial policy training. This represents a scenario where RL-trained policies for ACs are deployed to the real world and their weights are fixed in order to train any adversarial agent for testing. At this point, the Markov game consisting of two players can be treated as one-player MDP problem, since the victim policy $\pi_T$ is held fixed.

	The goal of the adversarial player is to learn a policy $\pi_\alpha$ maximizing the sum of discounted rewards:
	
	\[
	\pi_\alpha = \sum_{t=0}^{\infty} \gamma^t R_\alpha (s^{(t)},a_\alpha^{(t)},s^{(t+1)})
	\]
	
	where $a_\alpha \sim \pi_\alpha (. | s^{(t)})$ are actions sampled from the adversary policy and $ s^{(t+1)} \sim P_\alpha (s^{(t)},a_\alpha^{(t)})$ is the next state given the transition probability. Since the current problem is scoped as a model-free approach, the MDP dynamic model $P_\alpha$ is unknown.

	When the adversarial policy is trained, we use it to find uncommon behavior patterns for the victim's players by adding natural observations (see Section \ref{sec:Setup}) for the victim DRL policies $\pi_T$. 
	
	\subsection{Improving the Robustness of AC Policies by Retraining}
	Once we observe the effectiveness of the adversary in finding failure test cases for the ACs, we retrain the victim models by unfreezing their weight parameters, while keeping the trained adversarial player as part of the training environment. This leads to improved \textit{robustness} which in terms of DRL performance is its resistance towards out-of-distribution inputs and adversarial attacks~\cite{drenkow2021robustness}. Thus, the goal of the autonomous agents $\{\pi_{T1},\pi_{T2}\}$ is to maximize the sum of the discounted reward independently, that is:

	\[
	\pi_{T1} = \sum_{t=0}^{\infty} \gamma^t R_{T1} (s^{(t)},a_{T1}^{(t)},s^{(t+1)})
	\] 
	\[
	\pi_{T2} = \sum_{t=0}^{\infty} \gamma^t R_{T2} (s^{(t)},a_{T2}^{(t)},s^{(t+1)})
	\]

	%=====================================================================================%
	
	%=====================================================================================%
	
	\section{MAD-ARL Framework}\label{sec:Setup}
	
\begin{figure*}[!htbp]
		\centering
		
		\includegraphics[width = 0.9\textwidth,height=0.4\textheight]{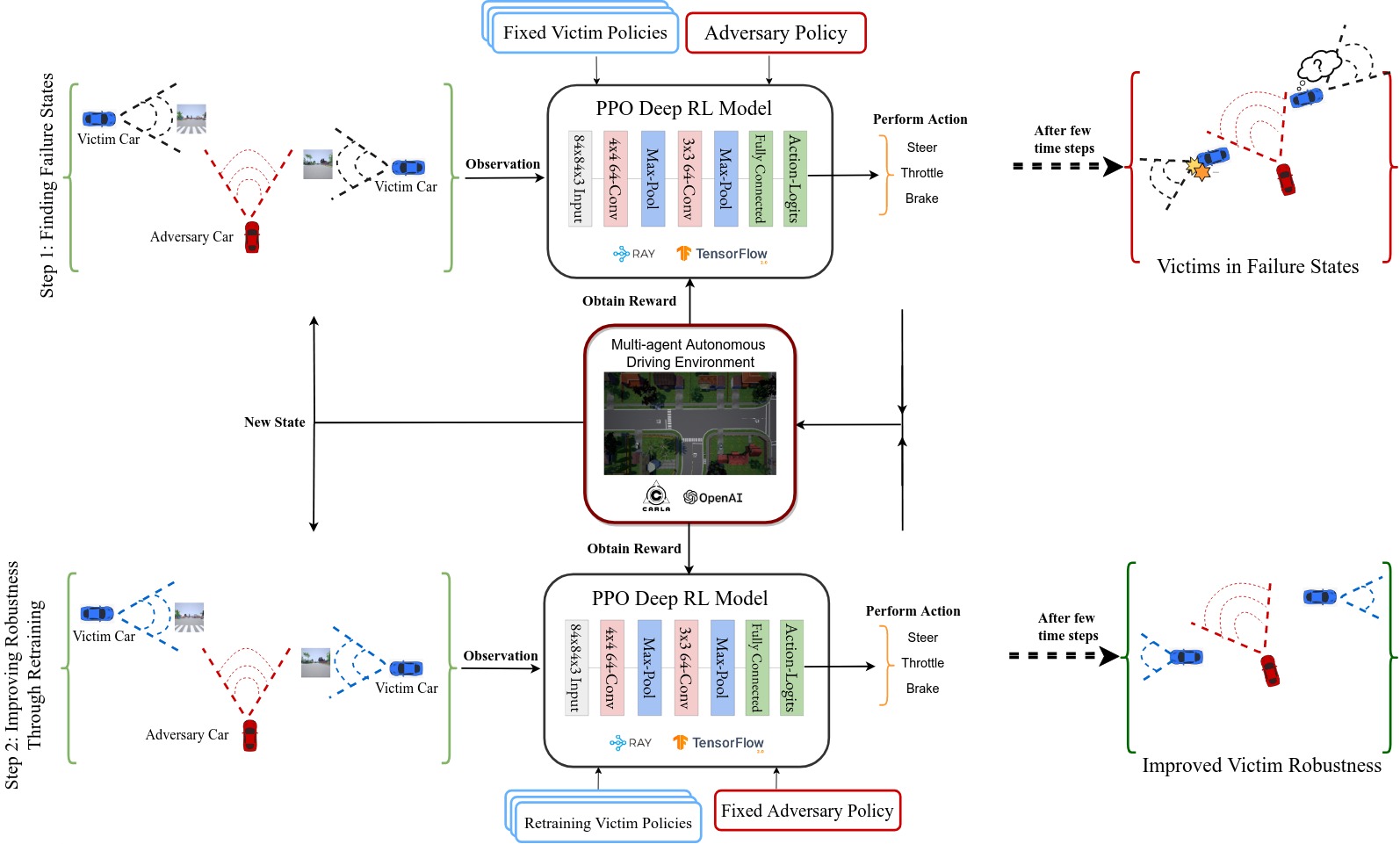}
		\caption{Illustration of MAD-ARL framework for improving the robustness of AC driving policies in a multi-agent environment. 
			Each agent receives an input image of 84x84x3 which is passed to a PPO-based DRL model. The actions are selected at the output layer of every agent and are performed in the next time step of the simulation in order to obtain reward and a new observation state. 
			Top row of the diagram displays the first step where we find failure scenarios of the victim policies, whereas the bottom diagram shows the second step that involves retraining of victims. Both steps of the framework are performed in an urban driving simulated environment.}
		%	\caption{End-to-end architecture with deep reinforcement learning model for autonomous and adversary agents. Each agent receives an input image of 84x84x3 which is passed to PPO based deep RL model. The actions are selected at the output layer of every agent and are performed in the next time step of the simulation in order to obtain reward and new observation state.}
		\label{fig:Architecture}
	\end{figure*}

	In this section, we present the proposed framework and the methodology for improving the robustness of AC driving policies in a multi-agent environment. An overview of the two-step methodology using the MAD-ARL framework is illustrated in Figure~\ref{fig:Architecture}.

 In the MAD-ARL framework, we consider two victim driving agents and one adversary driving agent. An \textit{\textbf{agent}} is an entity that is able to observe the environment and perform actions in order to make an intelligent decision in the given environment. % %Their architecture is discussed next.
The observations of our multi-agent driving environment for the victim agents are manipulated by adding an adversarial agent in the environment. The adversarial agent is trained to take actions such to create observations that appear \textit{natural} to the victim agents while being adversarial in nature. As an example, the adversary is learning to steer offroad most of the time while crossing the intersection. Such unusual behavior will act as an adversarial noise to the visual observations of victim ACs. The same framework is utilized to retrain the weights of the victims against the adversary in order to increase robustness and improve driving performance by lowering the number of collisions and offroad steering failures.

\subsection{Proximal Policy Optimization}\label{sec:PPO}
Our AC agents use Proximal Policy Optimization (PPO)~\cite{R55} as a policy gradient method to learn a driving policy by encountering a simulated environment in each training episode. The PPO helps perform on-policy learning within simulation instead of a dataset (replay buffer) type of learning. It also helps focus on policy updates with stability while learning over a change in data distributions, as well as address a large hyperparameter initializing space.

	The details of the hyperparameters selected for the training of the victim and adversary driving agents are given in Table~\ref{tab:HyperparametersPPO}.
	
	\begin{table}[htbp]
		\caption{Hyperparameters for the PPO DRL model.} \label{tab:HyperparametersPPO} 
		
		\begin{center}

			\resizebox{!}{2.0cm}{

				\begin{tabular}{lll}
					
					\toprule
					
					\textbf{Stage} & \textbf{Hyperparameter} & \textbf{Value} \\ \midrule

					\multirow{3}{*}{Gathering Experience} & Minibatch Range & 64 \\  
					
					& Epochs per Minibatch & 8 \\ 
					
					& Batch Mode & Complete Episodes \\ 
					
					\midrule
					\multirow{3}{*}{Updating Policy} & Discount factor ($\gamma$) & 0.99 \\ 
					
					& Clipping ($\epsilon$)  & 0.3 \\ 
					& KL Target  & 0.03 \\ 
					& KL Initialization & 0.3 \\ \midrule
					
					\multirow{2}{*}{Other Hyparameters } & Value Loss Coefficient   & 1.0 \\ 
					& Entropy Regularizer & 0.01 \\  \bottomrule
					
				\end{tabular}

			}

		\end{center}

	\end{table}

	\subsection{Deep Neural Network Model}\label{sec:mnih15}
	
	The summary of the DRL architecture, including the input, hidden, and output layer is displayed in Figure~\ref{fig:Architecture}. The input state $S \subset \mathbb{R} $ of our DRL algorithm receives a partial observation of 84x84x3 dimension images from the camera sensors. Cameras are mounted in front of each driving model which provides feeds as an input state observation to the autonomous and adversary cars model at each step of the simulation. The 3-dimensional input images are passed through convolutions and hidden layers to reach the output layer for control commands. 
	
	%	\begin{figure}[!htbp]
	%		\centering
	%		
	%		\includegraphics[width = 0.49\textwidth,height=0.23\textheight]{DeepRL.pdf}
	%		\caption{End-to-end architecture with deep reinforcement learning model for autonomous and adversary agents. Each agent receives an input image of 84x84x3 which is passed to PPO based deep RL model. The actions are selected at the output layer of every agent and are performed in the next time step of the simulation in order to obtain reward and new observation state.}
	%		\label{fig:deepRL}
	%	\end{figure}
	
	At the output layer, we have nine discrete values as the action space which are used by each driving agent to make control decisions. All of the discrete actions can be summed into three main actions: Steer, Throttle, and Brake. 
	
	\subsection{Reward Functions}\label{sec:Reward}
	
	Each agent is following MDP described in the MAD-ARL formulation, and therefore, at each time step, the driving models collect trajectories of $(S, R, A)$. $R$ is the reward gained in return for the actions chosen by the driving car's policy function, given the input observations.
	
	\subsubsection{Victim Reward Function}
	
	The victim policies are trained as driving ACs with the goal to safely reach as close to the desired destination as possible. The victim agents reward can be described as:
	\vspace*{1.5em}
	%\begin{align*} 
	%	R_{Victim} = (D_{t-1} - D_{t}) + (F_t)/10 -100.0(CV_t + CO_t) \\
	%	- 0.5(IO_t + IOL_t) + \beta
	%\end{align*}
	\begin{center}
		
		\resizebox{!}{0.95cm}{
			
			\begin{minipage}{0.5\columnwidth}
				
				\begin{align*}
				\label{eq:ab_crypto}
				\hspace*{-1.5em}
				R_{Victim}  = 
				\tikzmarknode{n}{\highlight{purple}{$(D_{t-1} - D_{t})$}} + \tikzmarknode{mi}{\highlight{NavyBlue}{$(F_t)$}}/10 -100.0\tikzmarknode{lij}{\highlight{Bittersweet}{$(CV_t + CO_t)$}} \\ - 0.5{\tikzmarknode{lmax}{\highlight{OliveGreen}{$(IO_t + IOL_t)$}}}
				+ \beta
				\end{align*}
				
				\vspace{0.8\baselineskip}
				\begin{tikzpicture}[overlay,remember picture,>=stealth,nodes={align=left,inner ysep=1pt},<-]
				% For "N"
				\path (n.north) ++ (0,1.0em) node[anchor=south east,color=Plum!85] (ntext){\textsf{\footnotesize Distance}};
				\draw [color=Plum](n.north) |- ([xshift=-0.3ex,color=Plum]ntext.south west);
				% For "M_i"
				\path (mi.north) ++ (0,2.0em) node[anchor=north west,color=NavyBlue!85] (mitext){\textsf{\footnotesize Forward Speed}};
				\draw [color=NavyBlue](mi.north) |- ([xshift=-0.3ex,color=NavyBlue]mitext.south east);
				% For "l_i^j"
				\path (lij.north) ++ (0,1.8em) node[anchor=north west,color=Bittersweet!85] (lijtext){\textsf{\footnotesize Collision}};
				\draw [color=Bittersweet](lij.north) |- ([xshift=-0.3ex,color=Bittersweet]lijtext.south east);
				% For "l_i^max"
				\path (lmax.north) ++ (-2.7,-2.0em) node[anchor=north west,color=xkcdHunterGreen!85] (lmaxtext){\textsf{\footnotesize Offroad steering}};
				\draw [color=xkcdHunterGreen](lmax.south) |- ([xshift=-0.3ex,color=xkcdHunterGreen]lmaxtext.south west);
				\end{tikzpicture}
				
			\end{minipage}
		}
	\end{center}
	where $D$ is the distance covered, $F$ is the forward speed of the agent, $CV$ and $CO$ are the boolean values telling whether there is any collision with other vehicles and environment objects, and $IO$ along with $IOL$ refers to driving offroad at the intersections and outside the desired lane represented as boolean. At the end of the equation is a constant $\beta$ used to encourage driving in a desired ground truth lane. From the above equation, it is clear that the victim driving policies are sensitive toward any offroad steering errors and collisions.
	
	\subsubsection{Adversary Reward Function}
	
	We are defining two different types of reward functions for the adversarial player to see which one performs better in testing and retraining the policies of the victim ACs. The two adversarial reward functions associated with the policies $\pi_{\alpha1}$ and $\pi_{\alpha2}$ are named  $R_{collision}$ and $R_{offroad}$. The motivation behind two different reward based adversaries is to show that the adversary with no collision and minimal offroad steering reward function is enough to create effective adversarial actions than the collision-focused adversarial agent.

	$R_{collision}$ aims to maximize the rate of collision and offroad steering during the adversarial training. $R_{collision}$ is formulated as:
	%\begin{align*} 
	%	R_{collision} = (D_{t-1} - D_{t}) + (F_t)/10 +5.0(CV_t + CO_t) \\
	%	+ 0.05(IO_t + IOL_t) + \beta
	%\end{align*}
	
	\begin{center}
		
		\resizebox{!}{0.43cm}{
			\begin{minipage}{0.2\columnwidth}
				\begin{align*}
				%				\label{eq:ab_crypto2}
				\hspace*{-0.5em}
				R_{Collision}  = 
				(D_{t-1} - D_{t}) + (F_t)/10
				\overbrace{\tikzmarknode{lij}{+5.0\highlight{Bittersweet}{$(CV_t + CO_t)$}} + 0.05{\tikzmarknode{lmax}{\highlight{OliveGreen}{$(IO_t + IOL_t)$}}}}^{\substack{\text{\sf \footnotesize \textcolor{red!85}{Collision and Offroad steering in $R_{Collision}$ Adversary}} } }
				\end{align*}
			\end{minipage}
		}
	\end{center}

   \vspace*{0.2em}
   $R_{offroad}$ on the other hand aims to maximize the rate of offroad steering, and is thus formulated as:
	
	%\begin{align*} 
	%	R_{offroad} = (D_{t-1} - D_{t}) + (F_t)/10  + 0.05(IO_t + IOL_t) \\ + \beta
	%\end{align*}
	\begin{center}
		\resizebox{!}{0.72cm}{
			\begin{minipage}{0.2\columnwidth}
				\begin{align*}
				%				\label{eq:ab_crypto2}
				R_{Offroad}  = 
				(D_{t-1} - D_{t}) + (F_t)/10
				%			\color{purple}
				\overbrace{+ 0.05{\tikzmarknode{lmax}{\highlight{OliveGreen}{$(IO_t + IOL_t)$}}}}^{\substack{\text{\sf \footnotesize \textcolor{red!85}{Offroad steering only in 
						}} \\ \text{\sf \footnotesize \textcolor{red!85}{$R_{Offroad}$ Adversary.
				}} } }
				\end{align*}
			\end{minipage}
		}
	\end{center}
	
	\subsection{Hyperparameters}\label{sec:Hyperparameters}
	
	The hyperparameters used in different phases of training of all ACs are shown in Table~\ref{tab:Hyperparameters}. %The table displays values of the most important hyperparameters used during the training of autonomous and adversarial cars.
	During the testing phase, explained in Section~\ref{sec:Results}, we run 50 total episodes, each having 2000 simulation steps per driving agent. 
	
	\begin{table}[htbp]
		\caption{Hyperparameters for the training of the baseline victim AC models, the adversarial model, and retrained victim AC models} \label{tab:Hyperparameters} 
		
		\begin{center}
			\resizebox{!}{1.15cm}{
				\begin{tabular}{llll}
					
					\toprule
					
					\textbf{Hyperparameter} & \textbf{Baseline} & \textbf{Adversarial}& \textbf{Retrained Victim} \\ \midrule

					Total Training Steps & 300672 & 57728 & 133888 \\ 
					
					Total Training Episodes & 610 & 101 & 306 \\ 
					
					Learning Rate & 0.0006 & 0.0006 & 0.0006 \\ 
					
					Batch Size & 128 & 128 & 128 \\ 
					Optimizer	& Adam~\cite{R59} & Adam &  Adam  \\ \bottomrule
					
				\end{tabular}
			}
		\end{center}
	\end{table}
	
	The details of the hyperparameters for all the AC and adversary agents are provided in the GitHub repository~\footnote{https://github.com/T3AS/MAD-ARL}.
	
	%=====================================================================================%
	\section{Experiments}\label{sec:Experimental_Details}
	
	The experiments aim to demonstrate the effectiveness of the proposed framework for testing and improving driving policies in a multi-agent car environment.   %In this section, we describe the details for both and training phase as well as the setup environment. 
	To this end, first, we train a single adversarial driving agent against one victim AC agent to test more than one victim AC driving policy. The purpose of the adversary is to expose errors in the driving policies of the AC agents, such as the inability to avoid collisions and offroad steering accidents. Later we retrain the AC agents using the adversarial inputs and evaluate how much their driving policies improved compared to their baseline performance. %We use two different adversarial reward functions to compare the results of victim AC against each adversarial reward base policy. 
	
	Specifically, the research questions aim to evaluate:
	
	\textbf{RQ1:} How effective is the adversarial driving policy in finding failure driving scenarios in victim ACs?
	
	\textbf{RQ2:} Does retraining the victim ACs using the adversarial inputs improve the agent's performance in terms of reduced collisions and offroad steering errors?
	
	\subsection{Evaluation Metrics}\label{sec:Metrics}
	%We evaluate the performance of victim ACs using four metrics: \textbf{i}) the amount of collision with another vehicles, \textbf{ii}) the amount of collision with any other road objects, \textbf{iii}) the number of times an AC went offroad from its driving ground truth lane, and \textbf{iv}) the time it takes to have a first collision in an simulation episode.
	
	We evaluate the driving performance of victim ACs using the following metrics:
	\begin{itemize}
		\item $C_V$: rate of collision with another vehicles
		\item $C_R$: rate of collision with any other road objects
		\item $O_S$: rate of offroad steering from a ground truth driving lane
		\item $TTFC$: time it takes to have the first collision
	\end{itemize}
	
	To evaluate the effectiveness of the adversarial driving policy in finding failure driving scenarios in victim ACs, we compare the AC's baseline performance (no adversary in the environment) with its performance when driving in the environment with an adversary present.
	To evaluate the effectiveness of adversarial retraining as a strategy to improve the driving performance of victim ACs, we also compare the performance of victim AC's performance and retrained victim AC's performance when driving in the environment with an adversary present. % with its baseline (no adversary in the environment) as well as during adversarial testing when trained with an adversary present in the environment.
	%we compare...   %The evaluation results are shown in Tables~\ref{tab:RQ1_RQ2}.  

	\subsection{Experimental Setup}
	We use \textit{Town 3} scenario provided by the Python Carla API and Macad-gym~\cite{R52} in our partially-observable urban-based driving environment. This environment has three independent non-communicating agents spawned close to the T-intersection throughout the training and testing steps, where two are the victim ACs $T_1$ and $T_2$, and one is the adversarial agent $\alpha$. The choice of T-intersection as a driving scenario is based on its higher complexity for an AC agent to handle, as the adversarial agent can be easily faced by victim policies during testing episodes.
	%	based on its higher complexity for 
	%	an AC agent in real life as well as in the autonomous driving research. 
	
	The goal of $T_1$ and $T_2$ is to drive straight across the intersection without errors, while $\alpha$ aims to take a left turn in the same driving situation. The starting and ending state locations of each driving agents are:
	\begin{itemize}
		\item $T_1$ start: [188, 59, 0.4], end: [167, 75.7, 0.13]
		\item $T_2$ start: [147.6, 62.6, 0.4], end: [191.2, 62.7, 0]
		\item $\alpha$ start: [170.5, 80, 0.4], end : [144, 59, 0]
		
	\end{itemize}
	where victim ACs strictly follow the mentioned coordinates as ground truth to improve their driving policies. On the contrary, the adversary player is less focused on reaching the desired destination and aims to deviate towards collision and offroad steering behavior.
	
	The sequence of the training and testing for victim and adversary agents are as follows.
	
	\begin{figure}[!htbp]
		\centering
		
		\includegraphics[width = 0.49\textwidth,height=0.23\textheight]{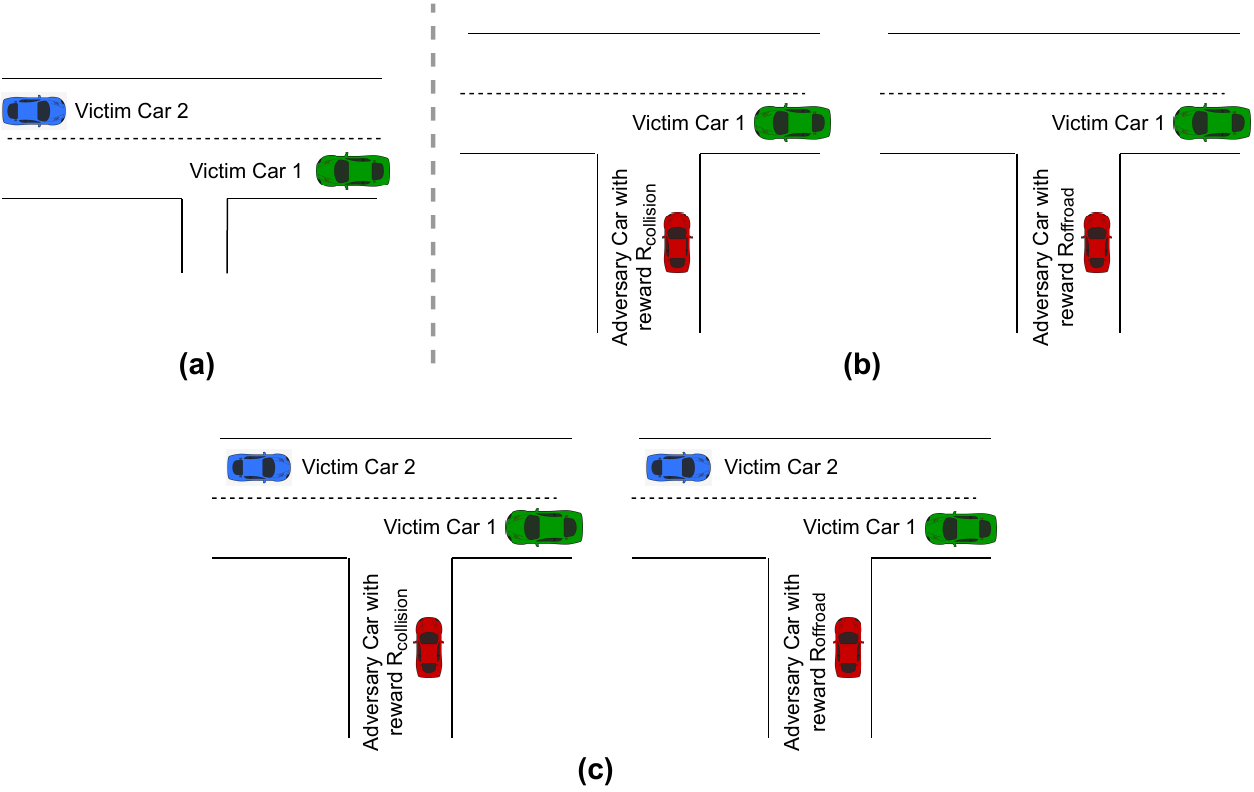}
		\caption{Illustration of the different phases of experimental setup. (a) shows the training phase of victim policies in the absence of an adversarial agent. (b) shows the training phase of both adversarial policies against one of the victim ACs. While (c)	illustrates Step 1 of the experiments where $R_{collision}$ and $R_{offroad}$ based adversaries are used to separately test against victim ACs. The same setup is used to retrain victim ACs for model evaluation in Step 2.}
		\label{fig:experimentalSteps}
	\end{figure} 
	
	\subsubsection{Training Victim AC Agents for Baseline}\label{sec:Training_Victim}
	%In the first step of our experiments, we use the setup described in Section~\ref{sec:Setup} to train two ACs. 
	We train both AC policies $\pi_{T1}, \pi_{T2}$ in a multi-agent environment with the absence of any adversarial policy as shown in Figure~\ref{fig:experimentalSteps}(a). After 610 episodes and 300672 steps mentioned in Section~\ref{sec:Hyperparameters}, we move towards our first testing phase to record the baseline performance of both autonomous cars.

	\subsubsection{Training Adversarial Agent}\label{sec:Training_Adversarial}
	% New changes
	Next, we introduce the adversarial driving agent. We train its policy $\pi_\alpha$ by providing the victim AC policy, keeping their weight parameters constant during the adversarial training phase as shown in Figure~\ref{fig:experimentalSteps}(b). The number of episodes for training the adversary agent is kept lower than the number of episodes assigned for the victim's baseline model training.

	We train the adversarial policy using two different adversarial reward functions, $R_{collision}$ and $R_{offroad}$, separately, as shown in Figure~\ref{fig:experimentalSteps}(b). This helps evaluate which adversarial agent is more effective in exposing errors in the victim ACs. We use 101 episodes to train the adversarial policy using both reward functions. 
	
\begin{figure}[!htbp]
		\centering
		
		\includegraphics[width = 0.495
		\textwidth,height=0.22\textheight]{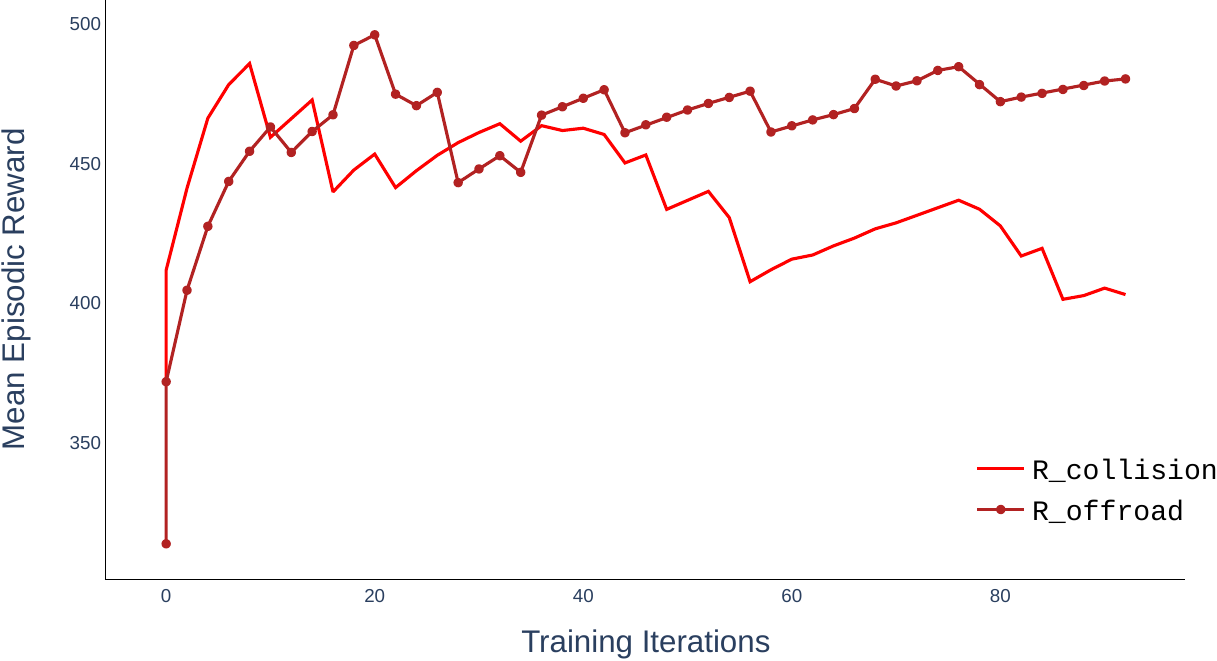}
		\caption{Scatter plot for showing the model training performance of the adversary agent using two reward functions. %Two adversarial policies are trained separately against victim ACs in a multi-agent driving environment. 
			Reward values are averaged during each adversary's training episode.}
		\label{fig:advrstrain}
	\end{figure}

	The performance of the adversarial policies that are individually trained is depicted in Figure~\ref{fig:advrstrain}. By training the DRL policies for 101 episodes, the adversarial policy $\pi_{\alpha2}$ trained on $R_{offroad}$ reward function converges faster than $\pi_{\alpha1}$ trained on $R_{collision}$. The policy $\pi_{\alpha2}$ gets to a stable mean episodic reward state after crossing half of the training steps, while $\pi_{\alpha1}$ tends to fluctuate throughout the training phase. %We can see that $R_{offroad}$ used in the adversary policy training performs slightly better than the $R_{collision}$. 
	Still, both reward-based adversarial policies lead the victim ACs into error states when tested.
	
	\subsubsection{Two-step Improvement of the Robustness of Victim ACs}

	\textbf{Step 1: Finding Failure States}\label{sec:Step1}
	We test the behavior and control decisions of both victim AC agents when exposed to the trained adversarial driving agent and compare the results with our baseline victim policies.
	The adversary is able to learn a generalized agent that is used to simultaneously attack both victim policies in a shared driving scenario as displayed in Figure~\ref{fig:experimentalSteps}(c). 
	Using the evaluation metrics described in Section~\ref{sec:Metrics}, we compare the driving behavior of the victim ACs as baselines against adversary agents. The results are described in Section~\ref{sec:Results}.
	
	\textbf{Step2: Retraining Victim ACs for Improved Robustness}\label{sec:Step2}
	Finally, we unfreeze the weights of the victim AC agents to retrain their end-to-end driving policies by keeping the adversarial agent in the same environment. %This time, both ACs train their end-to-end driving policy in a multi-agent scenario, but with the addition of a third driving adversary player which was missing in the Step 1. 
	Since there are two different adversarial reward-based policies involved, the retraining of the victim ACs is done twice separately, as depicted in Figure~\ref{fig:experimentalSteps}(c). After the retraining is done, we test the victim ACs to see how much they improved compared to their baseline performance. 
	% (collected in steps~\ref{sec:Step1} and~\ref{sec:Step2}). 

	\subsection{Simulation Setup}\label{sec:Simulation_Setup}
	We use \textit{RLlib}~\cite{R54} from Ray framework which is an open source project providing a very fine-tuned and scalable RL implementation interface. We also use \textit{Carla}~\cite{R51} urban driving simulation framework for training, testing, and validating ACs. For integrating Carla and Open AI's Gym toolkit in a multi-agent urban driving environment, we utilize open source platform \textit{Macad-gym}~\cite{R52}. We are also using \textit{Tensorflow}~\cite{R53} version 2.1.0 within the RLlib library for creating DRL based model architectures.
	%We use the following frameworks in the experiments:
	%
	%
	%\textit{RLlib}~\cite{R54} is a submodule of the Ray framework which is an open source project that provides a very fine-tuned and scalable RL implementation interface. Using RLlib, we can integrate existing deep RL algorithms, along with their hyperparameters, in our environment setup, while creating a multi-agent scenario with non-shared policy graphs. We are currently using version 0.8.4.
	%
	%
	%\textit{Carla}~\cite{R51} is an urban driving simulation framework, widely used for training, testing, and validating various autonomous driving cars in large accessibility of scenarios and map conditions. For our problem, we are currently using version 0.9.4.
	%
	%
	%\textit{Macad-gym}~\cite{R52} is an open source platform created by integrating Carla and Open AI's Gym toolkit in order to enable research opportunities in multi-agent urban driving environments. By taking the basic functionalities of macad-gym, we have modified the framework for our scope of work, although Carla's environment interface and parameters are kept as it is for the sake of reproducibility. The current stable version of macad-gym in use is 0.1.4.
	%
	%\textit{Tensorflow}~\cite{R53} is one of the leading frameworks used to create machine learning based algorithms. We are using version 2.1.0 within the RLlib library.
	
	%=====================================================================================%
	\section{Results \& Analysis}\label{sec:Results}
	In this section, we present and discuss experimental results, which are made available in the GitHub repository~\footnote{https://github.com/T3AS/MAD-ARL}.
	\begin{table*}[!htbp]
		
		\begin{center}
			\caption{Comparison of the behavior of victim ACs before and after adding adversarial car in the environment, and after retraining using adversarial policies, in terms of $C_V$, $C_R$, $O_S$, and $TTFC$ metrics, averaged across 50 episodes. Victim ACs have more collisions and offroad steering errors under the presence of an adversarial agent, compared with baseline victim models. Retraining victim ACs with adversarial inputs improves their driving policies. 
			} \label{tab:RQ1_RQ2}
			\resizebox{!}{1.2cm}{
				\begin{tabular}{l|ll|llll|llll}
					
					& \multicolumn{2}{|c}{Baseline}   & \multicolumn{4}{|c|}{After Adversarial Training
					} & \multicolumn{4}{|c}{After AC Retraining } \\
					&  &  & \multicolumn{2}{|c|}{$R_{collision}$}  & \multicolumn{2}{|c|}{$R_{offroad}$}  &  \multicolumn{2}{|c|}{$R_{collision}$}  &   \multicolumn{2}{|c}{$R_{offroad}$}   \\
					\midrule
					\multirow{4}{*}{}  & Victim 1 & Victim 2 &  Victim 1   &  Victim 2    &  Victim 1    &  Victim 2   &  Victim 1    &  Victim 2    &  Victim 1   & Victim 2   \\ 
					Collision with cars & 0.0 & 0.0 &  0.19468   & 0.0956    &  \color{red}{0.5484}   &  \color{red}{0.4048}  &  0.2563   &  0.3934 & \color{teal}{0.0831}  &  \color{teal}{0.0698}  \\
					Collision with other objects  & 0.0184  & 0.0398 &  0.0   &  0.1533   &  \color{red}{0.6465}   & \color{red}{0.278}   &  0.0   &  0.1912   & \color{teal}{0.0}    &  \color{teal}{0.0566}  \\
					Offroad steering error & 0.0929 & 0.2828 &  0.1645   &  0.3769   & \color{red}{0.9069}   & \color{red}{0.9747} &  0.0425   &  0.2112   &  \color{teal}{0.0358}   &  \color{teal}{0.1688} \\
					
					%\midrule
					%Time To Collision (timesteps) & 0 & 0 &  445   & 228    &  103   &  105  &  687   &  847 & 113  & 106 \\
					Time To First Collision (seconds)  & -  & - &  59.736   &  27.98   &  13.8292   & 14.5   &  92.1116   &  113.5248   & 15.1688    &  14.23 \\
					%					Time To First Collision (seconds)  & 24.3344  & 21.4688 &  49.8686   &  47.0468   &  21.163   & 23.7984   &  46.05584   &  47.2   & 80.416    &  84.13 \\
					
				\end{tabular}
			}		
		\end{center}
	\end{table*}
	\subsection{Effectiveness of Adversarial Driving Policy in Finding Failure Driving Scenarios in Victim ACs}
\begin{figure*}[!htbp]
		\centering
		
		\includegraphics[width = 1.0\textwidth,height=0.26\textheight]{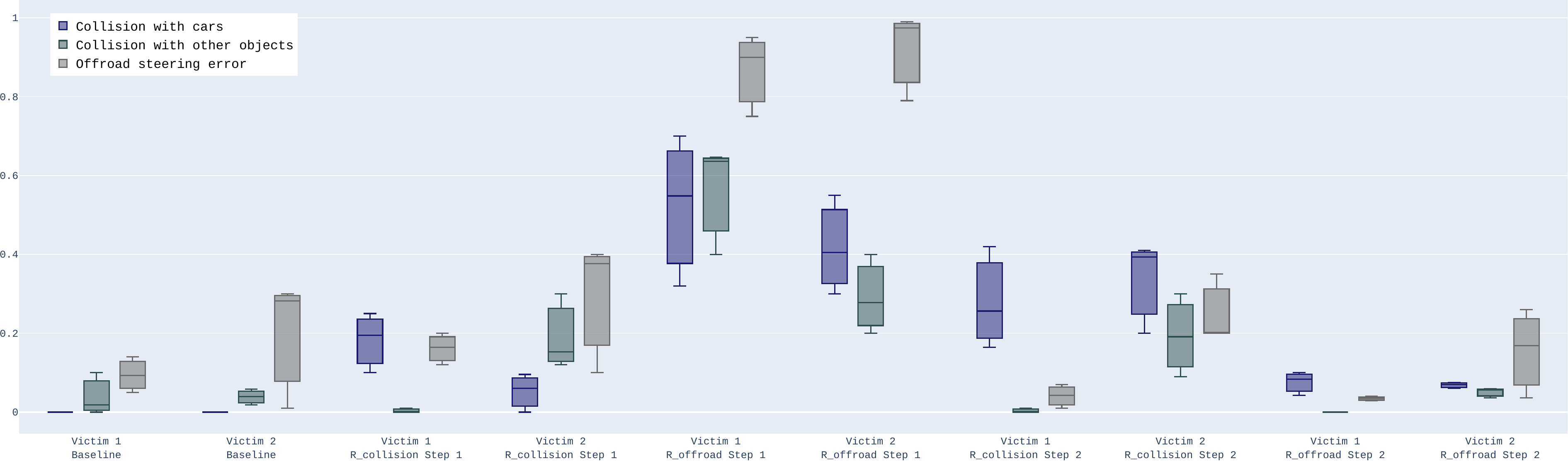}
		%\caption{Box plots for $C_V$, $C_R$, and $O_S$. The plot shows overall error rate of Victim 1 and Victim 2 within baseline, Step 1, and Step 2 of the experiments collected in testing episodes.}
		\caption{Overall performance of victim ACs before and after adding adversarial car in the environment, and after retraining using adversarial policies, in terms of $C_V$, $C_R$, and $O_S$ metrics.}
		\label{fig:box}
	\end{figure*}

	%We measure the effectiveness of the adversarial driving policy in exposing failures in victim ACs in terms of four metrics: $C_V$, $C_R$, $O_S$, and $TTFC$. We take an average among 50 episodic test runs, where in each episode we take 2000 simulation steps to evaluate the performance of victim agents. At the end of testing episodes, we compute the average percentage of errors. The results are presented in Table~\ref{tab:RQ1_RQ2}. For the first three metrics in the table $C_V$, $C_R$, $O_S$ victim policies having values closer to 0 are performing error-free driving, whereas values closer to 1 indicate a higher failure state of the victim policy. As for the fourth metric $TTFC$, the bottom two rows show the time it takes to detect the first collision in a testing episode, using timesteps and seconds respectively.  
	We measure the effectiveness of the adversarial driving policy in exposing failures in victim ACs in terms of four metrics: $C_V$, $C_R$, $O_S$, and $TTFC$. For $C_V$, $C_R$, and $O_S$, we calculate the percentage of error within each episode (as a value between 0 and 1). In each episode, we run 2000 simulation steps and at the end of 50 episodic test runs, we compute the average error rate for each metric across all the episodes.	The results are shown in Figure~\ref{fig:box} and the average error rate is presented in Table~\ref{tab:RQ1_RQ2}. For the first three metrics in the table $C_V$, $C_R$, $O_S$ victim policies having values closer to 0 are performing error-free driving, whereas values closer to 1 indicate a higher failure rate of the victim policy. As for the fourth metric $TTFC$, the bottom most row in the table shows the time in seconds it takes to detect the first collision in a testing episode.  
	
	%For conveying the results more clearly, we take an average among 5 episodic test runs, where in each episode we take 2000 simulation steps to evaluate the performance of victim agents. At the end of testing episodes, we compute the average percentage of errors. Victim policies having values closer to 0 are performing error-free driving, whereas values closer to 1 indicate a higher failure state of the victim policy.

	In the baseline scenario, both victims made no collision with each other during test episodes. Victim policies made uncertain decisions by creating collisions with footpaths and performing offroad steering errors. This is because we are testing the victim agents more than they have explored the environment in each episode. Victim 1, which is on the right side of the scenario, has a better baseline driving policy than Victim 2, as it has a lower rate of collision and offroad steering.
	%Consequently, Victim 2 is more negatively affected than Victim 1 overall once we add adversarial attacks in the environment.
	
	After introducing the adversarial policies to the environment ($R_{collision}$ and $R_{offroad}$), we see that the overall decision making process of both victim ACs is disturbed and their driving performance is decreased. Specifically, the rate of collision with other cars increased for both victims, as they ended up colliding with each other, as well as with the adversarial agent. Although both adversarial policies are finding the AC collision failure cases, $R_{offroad}$-based policy works better in this regard, which results in a higher rate of collision and offroad steering for both victim ACs (colored red in the table). With only offroad steering actions as adversarial action, $R_{offroad}$-based adversary forced victims into collision with each other in the T-intersection scenario. Also, due to early stopping after collision during $R_{collision}$-based policy, the rate of collision with other road objects has not been detected much. On the other hand, $R_{offroad}$-based adversarial policy is also able to find trajectories where victims end up hitting road objects after facing adversarial actions. Furthermore, both victims encountered more offroad steering errors. Victim ACs encountering $R_{offroad}$-based policy in a driving scenario ended up taking high percentage of offroad steering mistakes. Rather than attacking victims aggressively as by $R_{collision}$, $R_{offroad}$-based policy did it better by adding adversarial actions as natural adversarial observations produced in a shared driving environment.

	In terms of the time to first collision evaluation metric, there were no collisions in the baseline scenario for both victim ACs (denoted with dash in Table~\ref{tab:RQ1_RQ2}). After introducing the adversarial policies in the environment, collisions occurred after some seconds. Specifically, $R_{offroad}$-based policy drove victim ACs into collisions earlier than $R_{collision}$-based policy. %Looking it side by side with the other metrics, we see that for majority of the testing episodes in Step 1, victims against $R_{offroad}$-based policy were found colliding in earlier timesteps of a scenario.  
	%	The comparison of the baseline performance of victims versus adversarial agents in Step 1 is visualized in Figure~\ref{fig:TTC}.
	%	
	%	\begin{figure}[!htbp]
	%		\centering
	%		
	%		\includegraphics[width = 0.49\textwidth,height=0.22\textheight]{box.pdf}
	%		\caption{Box plot. Step 1 and Step 2 have results for both victims combined.}
	%		\label{fig:TTC}
	%	\end{figure}
	
	In summary, the results of the experiments demonstrate that introducing an adversarial policy to the environment is an effective strategy for finding failure driving scenarios in ACs.   %even though both $R_{collision}$ and $R_{offroad}$ based adversaries forced the victim ACs to drive off-road from their ground truth lane, 
	%$R_{collision}$-based policy discovering slightly more off-road steering cases due to its extreme adversarial driving actions.
	
	\subsection{Improving Victim ACs Performance by Retraining}
	
By retraining the victim ACs using inputs from $R_{collision}$- and $R_{offroad}$-based adversaries, we check whether the ACs' driving performance improved in terms of reduced collisions and offroad steering errors, compared to the stage before retraining. The evaluation results are shown in Table~\ref{tab:RQ1_RQ2}.  
Specifically, the rate of collision with cars and other objects decreased for the victims retrained using $R_{offroad}$-based adversary. The reason is that $R_{offroad}$-based adversary provides adversarial examples for victim ACs by maintaining collision-free distance, therefore helping victims to learn how to avoid collisions while crossing an intersection. $R_{collision}$ did not help much during the retraining process of the victim agents. It is mainly due to its collision-focused driving nature and thus the victims were unable to learn to avoid collisions with each other. Victims face collisions that are intentional by the adversary and most of the time they are unable to recover from such collisions in the episodic runs. Victim 2, being a weaker policy among the two ACs, also ended up colliding with road objects. Similarly, the number of offroad steering errors is reduced for the victims retrained using $R_{offroad}$ based adversarial policy since they slowed down after having collisions with other cars while retraining against $R_{collision}$, resulting in less space for improving the offroad steering behavior. 

Victims have neither seen such collision-focused drivers during baseline training, nor they are prepared for recovery steps when they face one in retraining. This is usually the case in training ACs in simulated or real-world datasets. The reason we have added $R_{collision}$ in our experiments is to show that it is not practical to add collision-focused driving cars around victims in a multi-agent environment. We need a better and more efficient framework where not only the RL-based AC victims are tested but also their robustness is improved.

Furthermore, the results show that the time to detect the first collision has been increased after retraining with the $R_{collision}$-based adversarial policy. For the $R_{offroad}$-based adversarial policy, even if the time to the first collision increased for only one victim AC, the overall rate of collision has decreased significantly for both ACs. This is because, after the first collision, the ACs were able to recover from failure states and continue with the safe driving behavior.

In summary, these results show that the retraining of victim ACs with adversarial policies helps in increasing the robustness of victims' driving performance. $R_{offroad}$-based adversarial policy once again proves to be more effective in adding better adversarial actions as observations to the driving scenes of the victims. Victims colliding in the testing episodes after retraining were able to recover and drive safe with minimal offroad steering errors. Overall, the results show that the $R_{offroad}$-based adversary is more effective in making ACs more robust. 

\begin{figure}[!ht]
		\subfloat[\label{genworkflow}][Victim ACs before retraining%\\ using $R_{collision}$
		]{%
			\includegraphics[, width=0.24\textwidth]{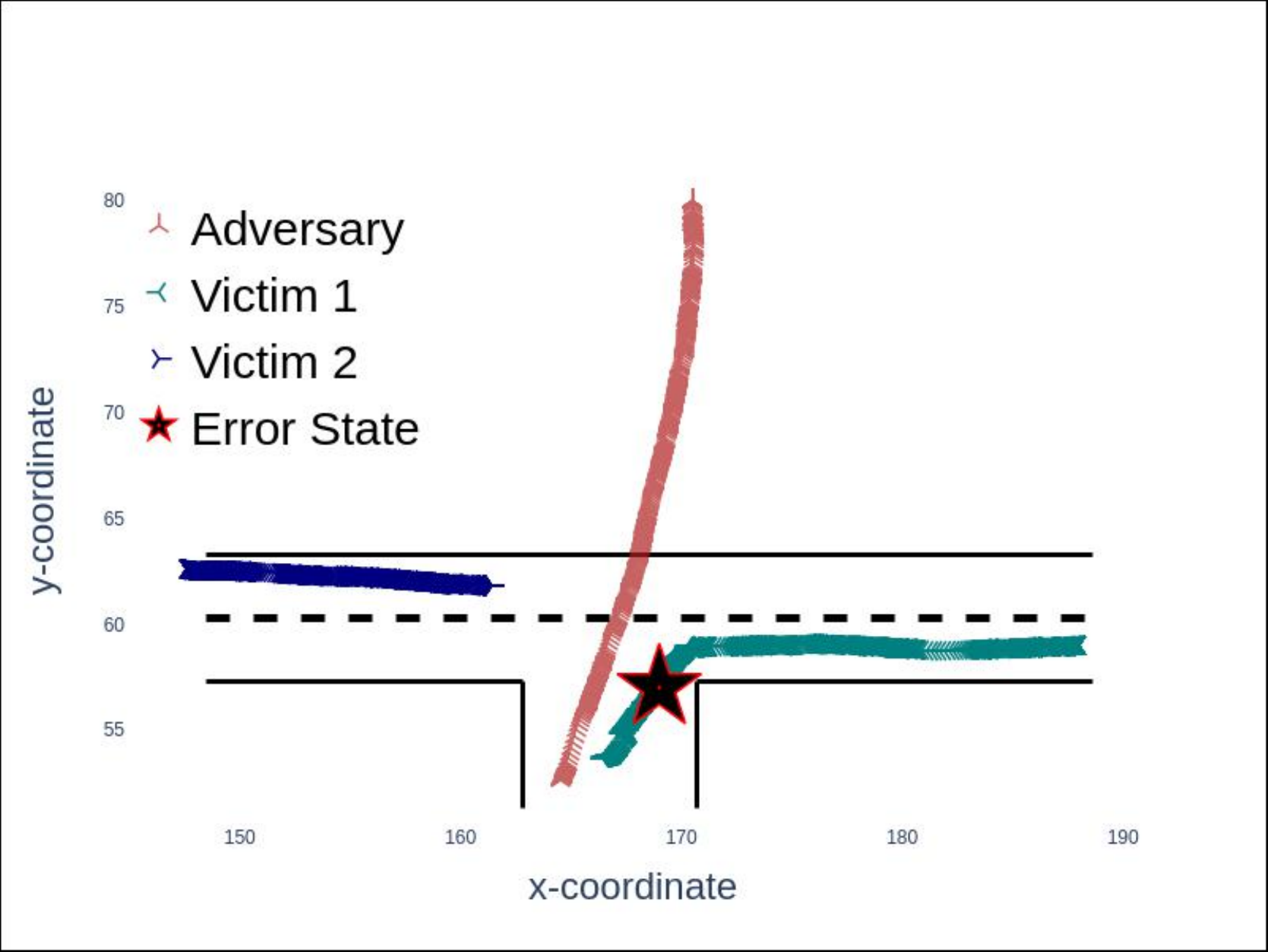}}
		%	\hspace{\fill}
		\subfloat[\label{pyramidprocess} ][Victim ACs before retraining%\\ using $R_{offroad}$ 
		]{%
			\includegraphics[width=0.24\textwidth]{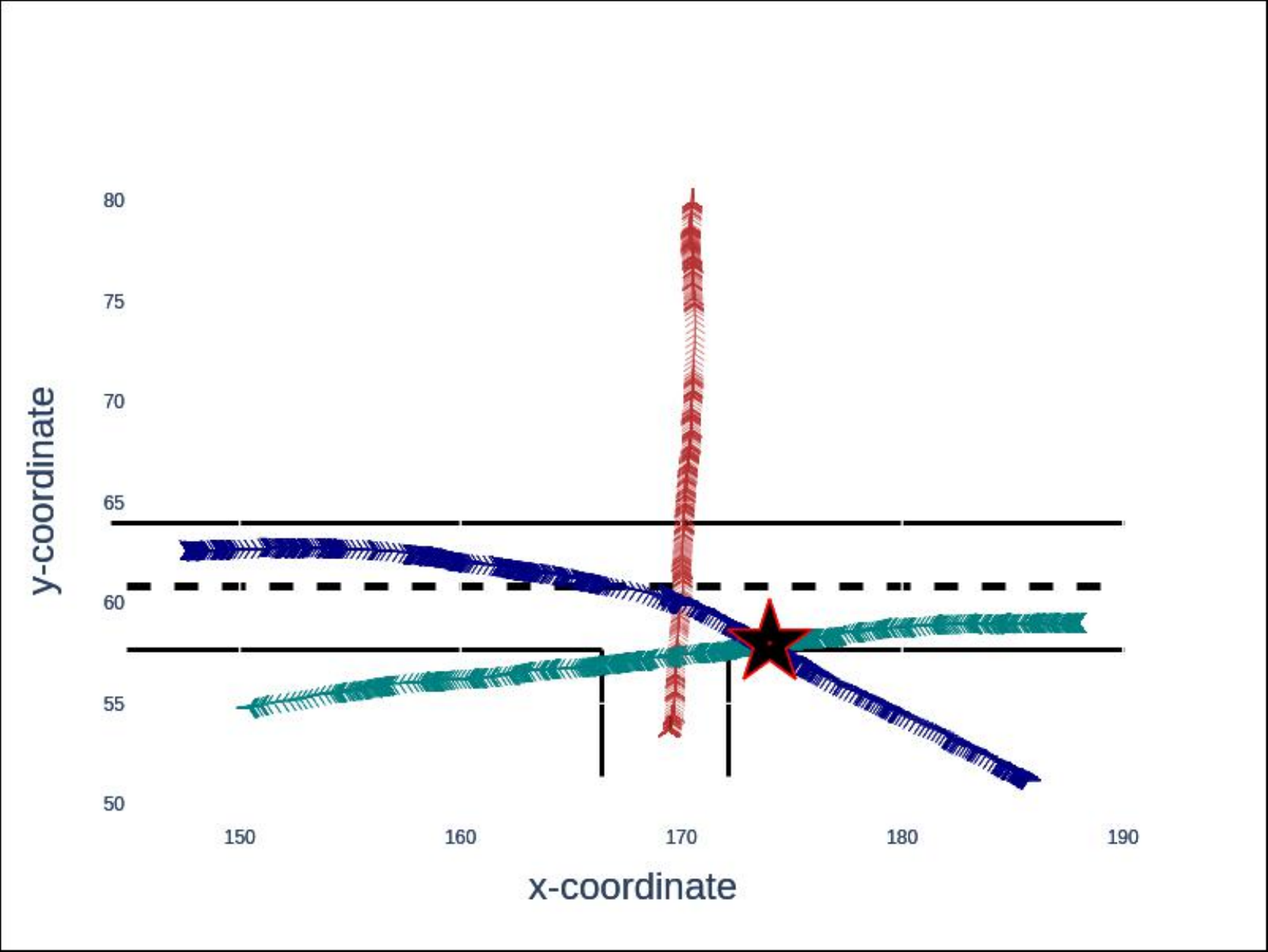}}
		\hspace{\fill}
		\subfloat[\label{mt-simtask}][Victim ACs after retraining %\\ using $R_{collision}$
		]{%
			\includegraphics[ width=0.24\textwidth]{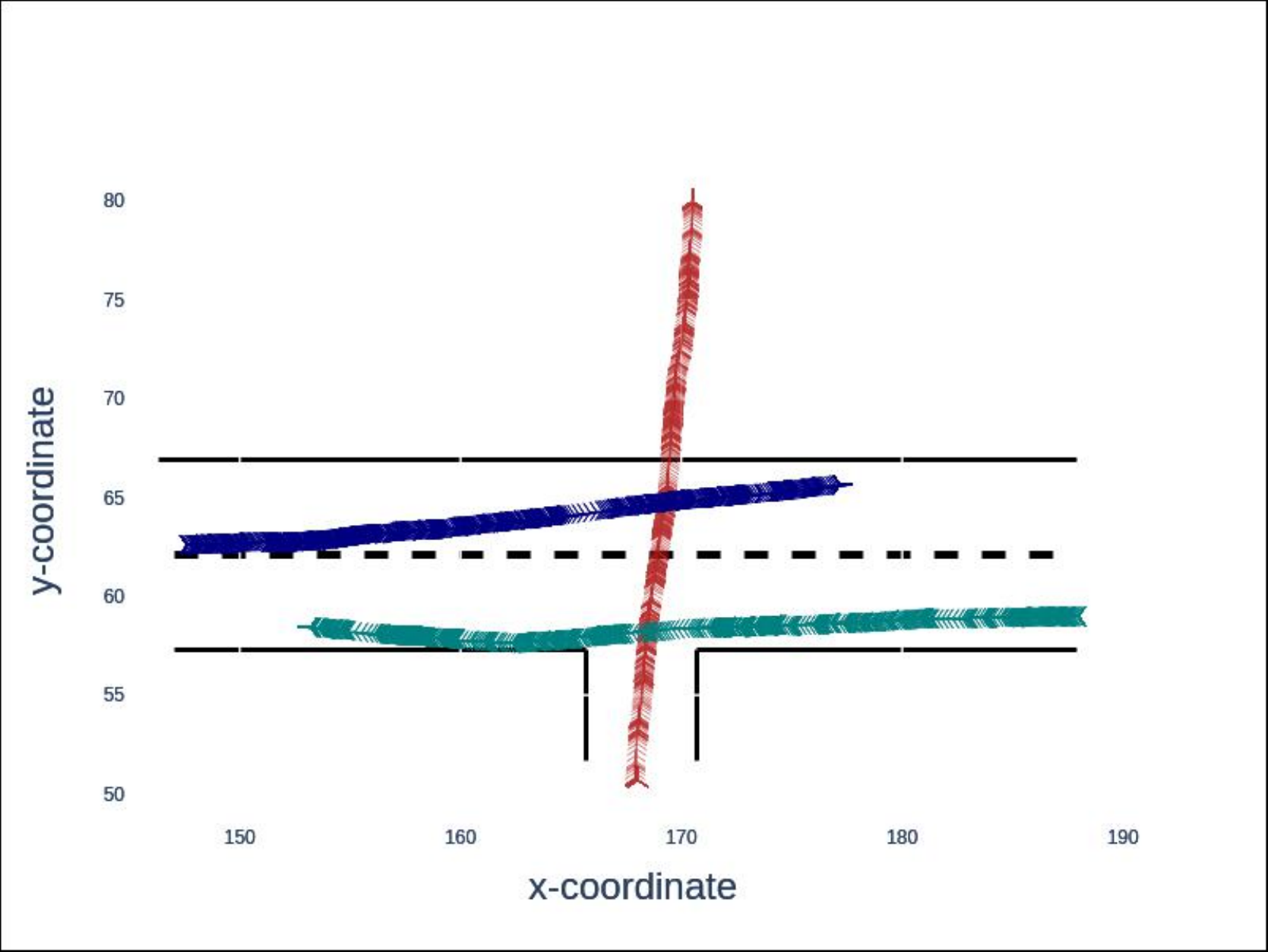}}
		\subfloat[\label{mt-simtask}][Victim ACs after retraining %\\ using $R_{offroad}$
		]{%
			\includegraphics[ width=0.24\textwidth]{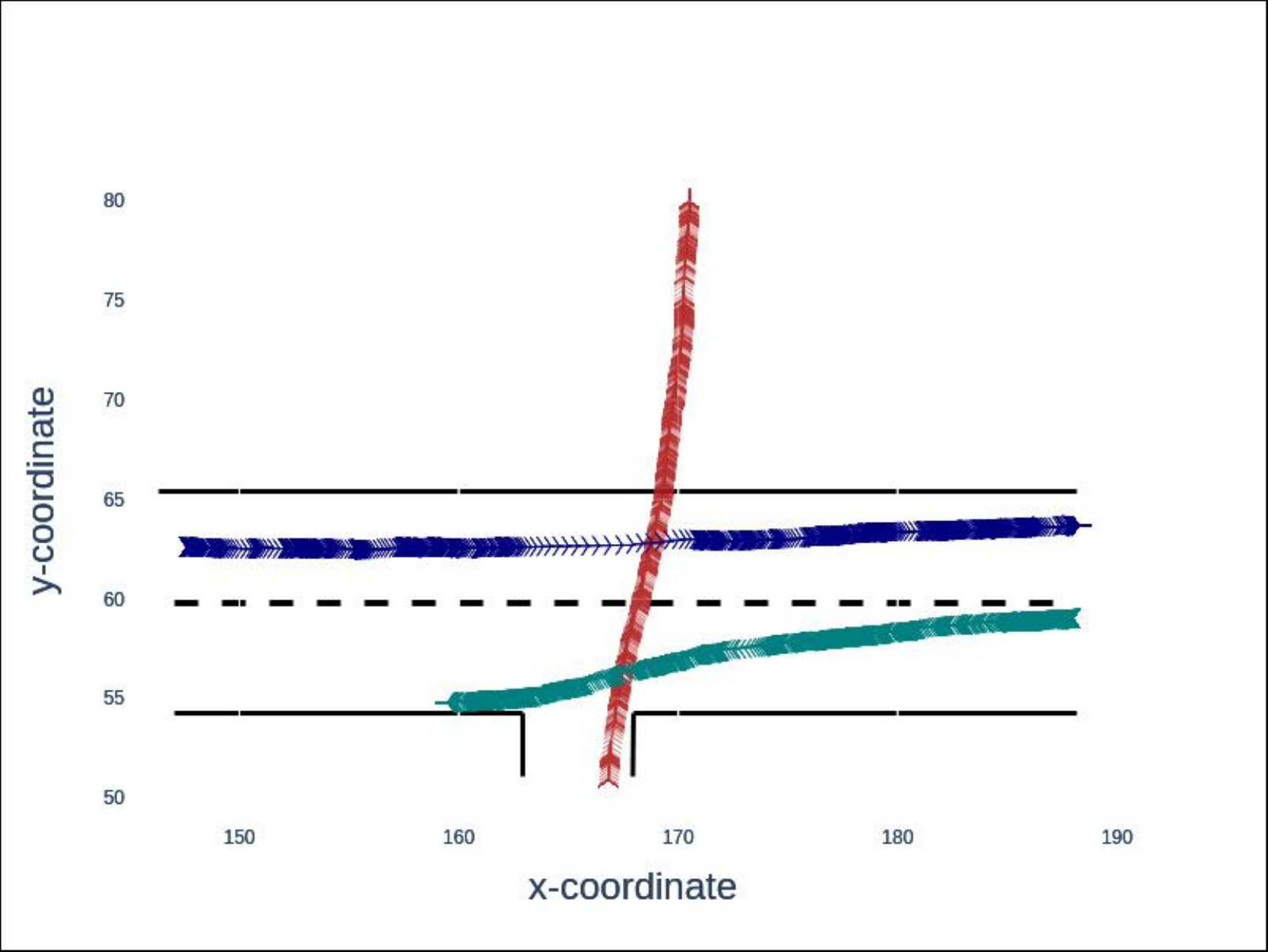}}\\
		
		\caption{\label{fig:location}2D visualization of the victim and adversary driving coordinates. (a) and (b) display two failure scenarios found while testing victim cars in the presence of an adversarial agent. (c) and (d) illustrate the same victim policies performing better once they are retrained with the adversarial policies. %car for retraining their deep RL policy. Left column figures are associated with $R_{collision}$ based adversary while the right column plots are related to $R_{offroad}$ based adversary. 
		}
	\end{figure}

	We visualize the driving performance of the victim AC agents before and after retraining in Figure~\ref{fig:location}. The figure shows a 2-dimensional aerial view of the victim ACs' driving coordinates. Plots (a) and (b) represent the performance when the victim agents are exposed to the adversary for the first time, while plots (c) and (d) represent the improvement in their driving policies as the result of retraining. The plots do not take a time factor into consideration, which is important to mention since any victim car overlapping with the adversary does not necessarily mean a collision state. Plot (a) depicts Victim 1 driving offroad without crossing the intersection, due to the adversarial agent. Plot (b) depicts a failure scenario where Victim 2 collides with the adversary and drives offroad. Error states in these two plots are marked with red stars. Plot (c) and (d) depict cases of improved (retrained) driving policies of the victim agents, who are now able to avoid collisions with cars and other road objects, as well as to stay in the driving lane while crossing the intersection.

	%=====================================================================================%
	\section{Conclusion \& Future Work}\label{sec:Conclusion}
	In this work, we propose a framework named MAD-ARL which is a multi-agent driving environment designed for improving the robustness of autonomous cars using adversarial driving models. ARL is trained against a victim player in order to find unwanted driving decisions of autonomous cars that are also trained on a DRL-based policy. By exposing the same adversarial car against the victim agents for retraining, the agents show improvements in their end-to-end decision driving controls, mainly in terms of fewer collisions and offroad steering errors compared to their originally trained (adversary-free) policies.
	
	%\textbf{Discussion}:
	%Deep RL research for ACs is mainly focused on driving in a single-agent stationary MDP environment, whereas MAD-ARL framework is based on POMDP formulation where every agent is following MDP by receiving partial observations in a competitive environment. In multi-agent non-stationary environments, each agent's transition probability and reward function depends on the actions of all the agents since they change every time with the actions performed by the agents.  The driving behavior of AC agents is therefore affected a lot when tested in a multi-agent scenario due to the non-stationary driving environment~\cite{Papoudakis2019DealingWN}.   
	
	\textbf{Future Work}: This work can be further extended to ACs operating in mass-traffic scenarios having more cars, pedestrians, and a traffic light network as part of the multi-agent environment. In such complex environments, mixed competitive ACs need to be tested in larger state space for finding edge cases using adversarial agents. Furthermore, we plan to investigate how retraining the adversarial agent affects the performance of victim autonomous cars. We also plan to explore and compare the robustness of different DRL algorithms used for autonomous driving research, when they are exposed to different types of adversaries. Also, we will extend current driving scenario with different training and testing episodic steps for evaluating the driving performance of RL-based models.
	
	%\section{Acknowledgment}
	%This work is supported by the Research Council of Norway through T3AS project.
	
	\bibliographystyle{IEEEtran}
	\bibliography{sample-base}
	
\end{document}